\documentclass{egpubl}
\usepackage{pg2022}
\usepackage{indentfirst}
\usepackage{amssymb}
\usepackage{amsmath}

\SpecialIssueSubmission    

\usepackage[T1]{fontenc}
\usepackage{dfadobe}  

\usepackage{cite}  
\BibtexOrBiblatex
\electronicVersion
\PrintedOrElectronic
\ifpdf \usepackage[pdftex]{graphicx} \pdfcompresslevel=9
\else \usepackage[dvips]{graphicx} \fi

\usepackage{egweblnk} 

\newcommand{\PHR}[1]{{\textcolor{black}{#1}}}

\title[SO(3)-Pose:SO(3)-Equivariance Learning for 6D Object Pose Estimation]%
      { SO(3)-Pose: SO(3)-Equivariance Learning for 6D Object Pose Estimation}

\author[Haoran Pan et al.]
{\parbox{\textwidth}{\centering Haoran Pan$^{1}$, Jun Zhou$^{2}$, Yuanpeng Liu$^{1}$, Xuequan Lu$^{3}$, Weiming Wang$^{4}$, Xuefeng Yan$^{1}$, Mingqiang Wei$^{1}$}\\
{\parbox{\textwidth}{\centering $^1$Nanjing University of Aeronautics and Astronautics\\ $^2$The Hong Kong Polytechnic University \\ $^3$Deakin University \\ $^4$Hong Kong Metropolitan University}
}
}

\begin{document}


\maketitle
\begin{abstract}
6D pose estimation of rigid objects from RGB-D images is crucial for object grasping and manipulation in robotics.
Although RGB channels and the depth (D) channel are often complementary, providing respectively the appearance and geometry information, it is still non-trivial how to fully benefit from the two cross-modal data. 
From the simple yet new observation, when an object rotates, its semantic label is invariant to the pose while its keypoint offset direction is variant to the pose. 
To this end, we present SO(3)-Pose, a new representation learning network to explore SO(3)-equivariant and SO(3)-invariant features from the depth channel for pose estimation. 
The SO(3)-invariant features facilitate to learn more distinctive representations for segmenting objects with similar appearance from RGB channels. The SO(3)-equivariant features communicate with RGB features to deduce the (missed) geometry for detecting keypoints of an object with the reflective surface from the depth channel. Unlike most of existing pose estimation methods, our SO(3)-Pose not only implements the information communication between the RGB and depth channels, but also naturally absorbs the SO(3)-equivariance geometry knowledge from depth images, leading to better appearance and geometry representation learning.
Comprehensive experiments show that our method achieves the state-of-the-art performance on three benchmarks.


\end{abstract}  

\section{Introduction}
6D pose estimation predicts a rigid transformation (i.e., 3D rotation and translation) from the 3D coordinate system of the object to the 3D coordinate system of the camera. Accurate 6D pose enables a robot to interact with target objects in the environment effectively~\cite{choi2012voting,du2021vision}.
Although recent years have witnessed a spurt of progress in 6D object pose estimation, it is still challenging and remains many open problems to solve, due to the heavy occlusions, cluttered backgrounds, and varying illuminations in the environment. 

\begin{figure}[t]
    \centering
    \includegraphics[width=1.0\linewidth]{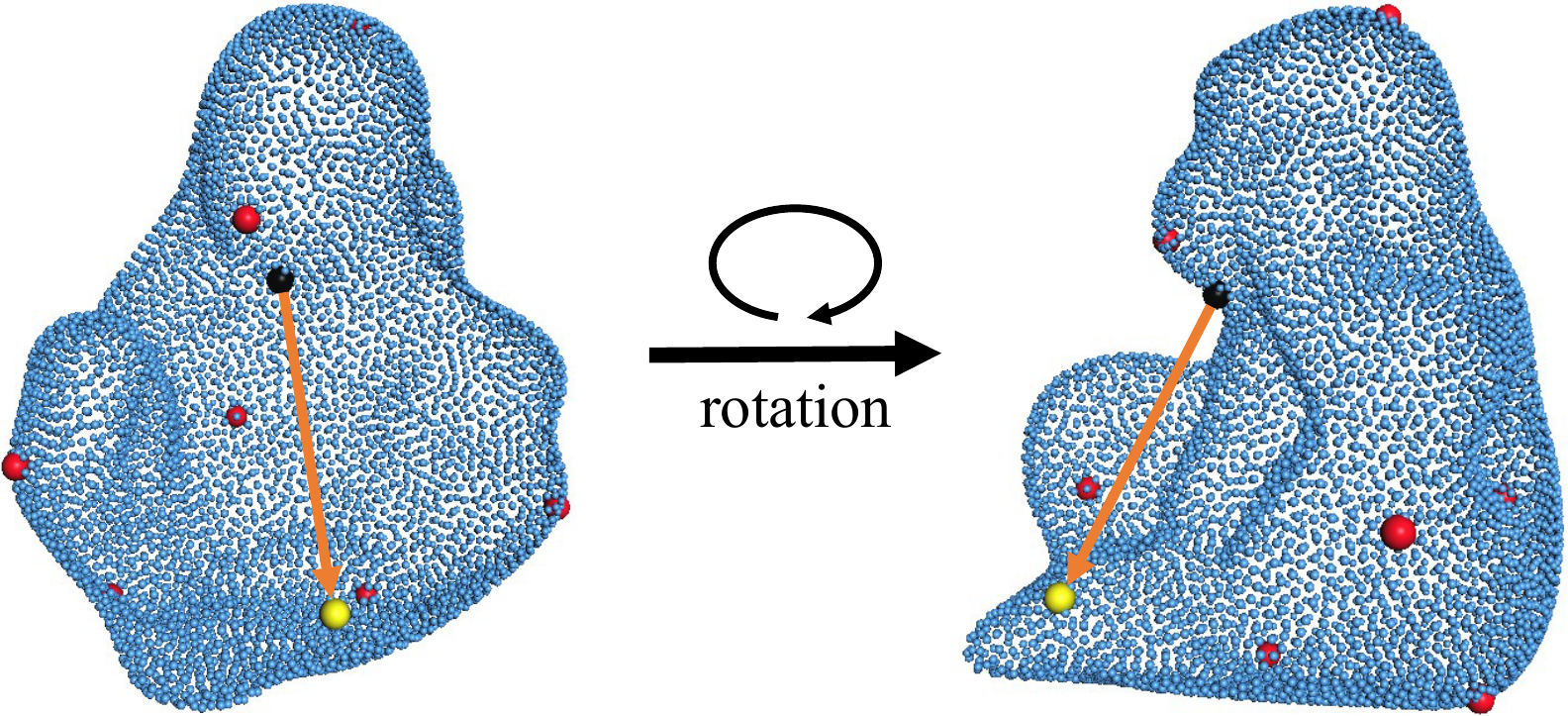}
    \caption{\PHR{
    Visualization of the SO(3) property. The red dots represent keypoints and the yellow dot is one of the keypoints. The black dot is an appearance point on the object. The semantic label (black dot) is invariant and the keypoint offset direction (the orange arrow between the black dot and the yellow dot) is variant when the object rotates.}
    }
    \label{fig:so3}
\end{figure}
\begin{figure*}[htbp]
    \centering
    \includegraphics[width=1.0\linewidth]{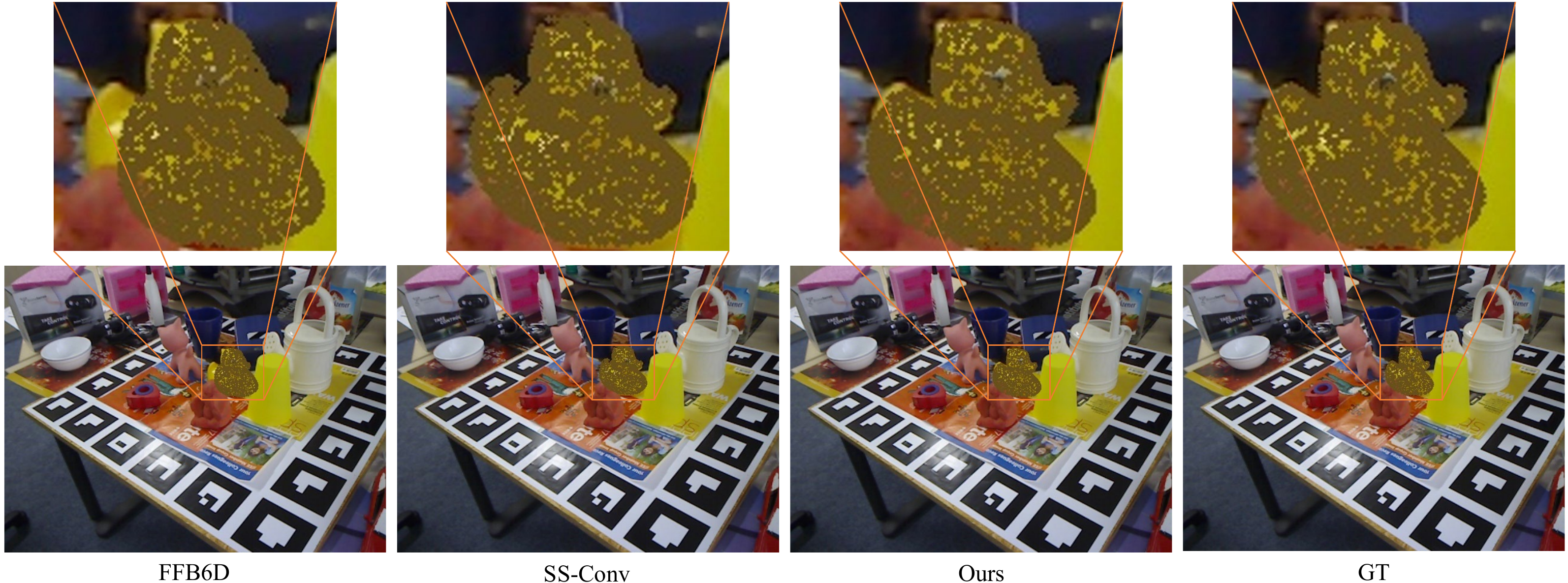}
    \caption{Comparison of our SO(3)-Pose with FFB6D \cite{he2021ffb6d} and SS-Conv \cite{lin2021sparse} for pose estimation on the LineMOD dataset. Our SO(3)-Pose predicts a more accurate 6D object (see the Duck) pose that best fits the duck in the image than the state-of-the-arts.
    Please note that, the object vertices in the object coordinate system are transformed by the predicted poses from individual methods to the camera coordinate system and then projected to the image with the camera intrinsic matrix. }
    \label{duckvis}
\end{figure*}

Recent methods mainly depend on learning-based techniques (e.g., CNN)~\cite{peng2020pvnet} for 6D pose estimation. That is, at the training stage, cutting-edging models learn from 3D objects and images that contain the objects in known 6D poses; and at the test stage, given a list of object instances and an image with the objects visible in it, the trained models infer 6D poses of the listed object instances. Initially, the inputs for these learning-based models are only RGB images. When encountering the cases of 1) similar, occluded or texture-free objects, 2) poor lighting conditions, and 3) low-contrast scenes, these models become difficult to learn distinctive representations from RGB images, due to 
lack of the geometry information. 
The recently popular 3D sensors, such as Microsoft
Kinect, Velodyne LiDAR, Intel RealSense, LiDAR scanner of Apple iPad Pro, can capture the real world into RGB-D images. The extra depth (D) information is promising to alleviate the pose estimation problem.

Given an RGB-D image, how to fully benefit from the two cross-modalities for better 6D pose estimation is still an open problem. The common way of handling the cross-modalities is to extract appearance features and geometry features separately by two-stream networks, e.g., DenseFusion~\cite{ChenWang2019DenseFusion6O}, and PVN3D~\cite{he2020pvn3d}. The appearance and geometry features from the two steams are fused and assigned to each pixel to realize the pose estimation. 
However, since the two streams seldom interact with each other to obtain the mutual gain, their performance surfers from degeneration in cases of e.g., objects with similar appearance or with reflective surfaces. FFB6D~\cite{he2021ffb6d} pioneers to communicate between the two streams by constructing bidirectional fusion modules. 
Thus, the two streams are encouraged to share the local-and-global complementary information from each stream for learning appearance and geometry representations.
Although FFB6D achieves the state-of-the-art performance, the extracted features are still simply fused to learn the semantic segmentation and keypoint offset, without exploring geometry domain knowledge. 
Furthermore, since the full flow bidirectional fusion operation is applied on each encoding and decoding layer of the two networks, the excessive interaction of the appearance and geometry features leads to colossal time consumption, which hinders its capability on real-time applications.
By contrast, our work delves into the geometric properties from the perspective of rigid transformations and turns to learn SO(3) equivalence and invariance instead of conventional geometric features, and therefore to estimate the 6D object pose effectively. Herein, SO(3)-equivariance/invariance is defined as a characteristic for feature mapping that the input and output have equivalent/invariant transformation effects with a given instance in the manifold space.

In this work, we propose SO(3)-Pose, a new representation learning network to explore SO(3)-equivariant and SO(3)-invariant features from point clouds for \PHR{instance-level object pose estimation which needs object mesh.} 
SO(3)-Pose adopts a two-stage strategy: 1) jointly segment the target objects from the RGB image with the help of SO(3)-invariant features and regress the keypoints of objects from the depth image with the help of SO(3)-equivariant features; and 2) given the keypoints, a PnP optimization problem is solved to produce the pose parameters. 
From our observation, the semantic label is SO(3)-invariant and the keypoint offset is SO(3)-equivariant (see Fig. \ref{fig:so3}).
To this end, we design an SO(3)-equivariant encoder to extract SO(3)-equivariant features from the point cloud, and meanwhile develop an equivariant-to-invariant layer (E2Ilayer) to make the SO(3)-equivariant features be the SO(3)-invariant features. The  SO(3)-equivariant features and  the RGB features are aggregated to localize keypoints, while the SO(3)-invariant features and the RGB features are aggregated to segment object instances.

Our method achieves the state-of-the-art performance (see Fig. \ref{duckvis}). In summary, our contributions are as follows:
\begin{itemize}
\item[1)] We propose a novel 6D object pose estimation network, which introduces SO(3)-equivariance for representation learning. To the best of our knowledge, this is the first work to introduce SO(3)-equivariance to pose estimation.
\item[2)] We design a new module termed  E2Ilayer, which effectively converts the SO(3)-equivariant features to the SO(3)-invariant features; we also design a novel loss, dubbed SO3loss, to guide the SO(3)-equivariance learning.

\item[3)] We show the SO(3)-equivariance benefits both tasks of semantic segmentation and keypoint detection by developing individual feature fusion modules.
\end{itemize}

\begin{figure*}[ht]
    
    \centering

    \includegraphics[width=1.0\textwidth]{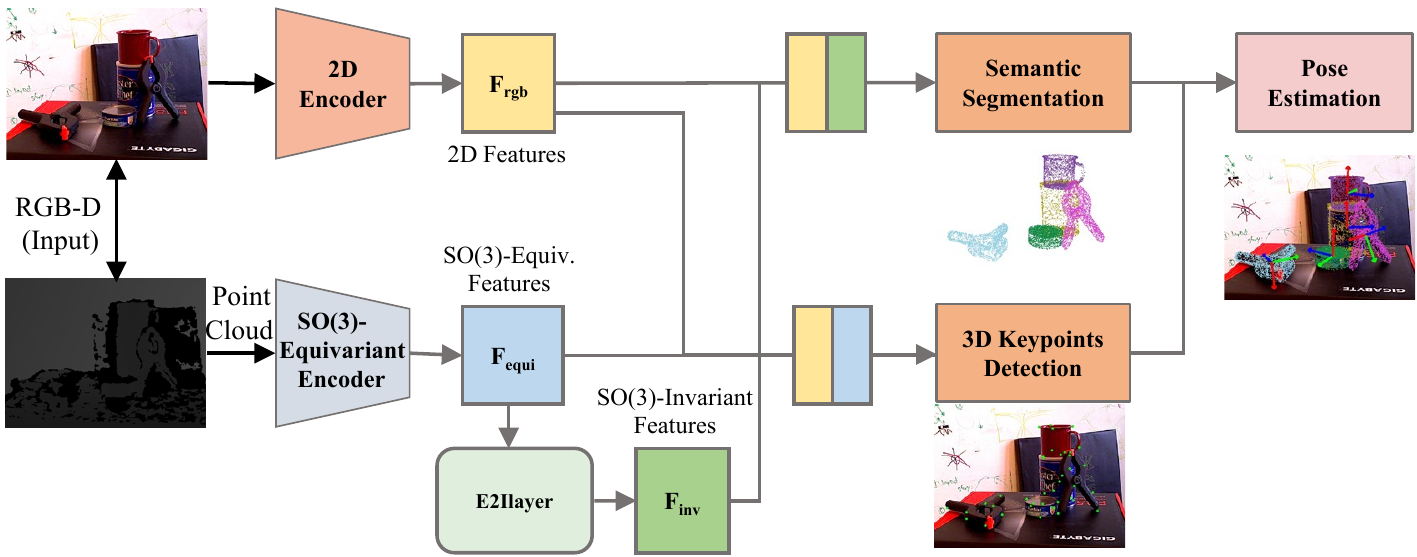}

    \caption{Overview of our SO(3)-Pose. SO(3)-equivariance is a geometric property in that a feature rotates the same degree and direction as the object rotates.
The SO(3)-invariant feature $F_{inv}$ is obtained from the SO(3)-equivariant feature $F_{equi}$ through the designed E2Ilayer. $F_{inv}$ and $F_{rgb}$ are fused and fed into a semantic segmentation module. $F_{equi}$ and $F_{rgb}$ are fused and fed into a 3D keypoint detection module to obtain per-object keypoints. Finally, the 6D pose parameters are fitted with the 3D-3D correspondence of keypoints. }
    \label{fig:overview}
\end{figure*}
\section{Related Work}
\subsection{Traditional approaches}
In the traditional setting, methods focus on holistic and local object shape representation learning. These methods can be roughly divided into two groups, template-based and descriptor-based, which are distinguished by how to utilize feature embedding and clustering for pose estimation. 

The first kind of method mainly aims at computing holistic shape description for each target model~\cite{du2019vision,hinterstoisser2012model,hodavn2015detection}. Specifically, the key technique of this type of method is template matching. These templates are generated by projecting a 3D model onto different image planes from various viewpoints, and each template has a pose parameter. In the inference phase, the final object pose will be recovered by the correlation coefficient between the query window and the template~\cite{zhang2017texture}. 
As an early attempt for this task, Hinterstoisser et al.~\cite{hinterstoisser2012model} proposed a classical framework, which integrates image gradients and surface normals for robust feature embedding. This scheme achieves distinguished holistic shape representation and significant detection performance.
Rios-Cabrera et al.~\cite{rios2013discriminatively} devised a real-time scalable approach based on LINE2D/LINEMOD~\cite{hinterstoisser2011gradient}. They also designed a novel strategy to distinguish templates by employing a cluster manner. 
This kind of method works well with texture-less objects, but the detection performance will be severely degraded when a high occlusion exists in real scenes. 

The second group of methods performs either 2D-3D or 3D-3D matching to establish correspondences in feature space, and the final pose parameters can be recovered by solving the PnP problem~\cite{lepetit2009epnp}.
The core insight of this class of methods is to extract robust feature points both on the image plane and 3D geometry surface~\cite{jiang2021review,ma2021image,hana2018comprehensive,li2021tutorial}, and employs feature description on each keypoint to obtain a series of matching points with feature similarity between them. 
One of the most representative approaches was introduced by Mur-Artal et al.~\cite{mur2015orb}, which proposed a novel 2D descriptor called ORB. The ORB features are firstly extracted on the given image and the pose parameters of keyframes are calculated by the constructed correspondences from the 2D pixels to 3D points. 
Correspondingly, when depth data is available, the 3D descriptors~\cite{salti2014shot,huang2021comprehensive,drost2010model,hinterstoisser2016going,vidal2018method,zhou2020bold3d} can be built on the model surfaces and the correspondence matrix will be filled with the matched 3D points. The pose estimation task turns into a rigid registration problem. Once the correspondence is established, the final pose can be solved by utilizing the SVD algorithm. 
In general, descriptor-based approaches have advantages in terms of computation complexity and robustness to partial occlusion, since only the local structural features of the model are highlighted and pushed into downstream calculation stages. However, due to the locality injection property of these methods, the proposed 3D descriptors are sensitive to the variation of the environment, e.g., illumination.

\subsection{Learning-based approaches}
Recently, with the prevailing attention of deep learning techniques, learning-based approaches have been deployed into many vision fields and reveal remarkable progress such as image classification~\cite{ali2021xcit}, object detection~\cite{dai2021up}, semantic segmentation~\cite{wang2021max} etc.
Similarly, learning-based methods can be roughly divided into two categories, i.e., holistic and semi-holistic approaches. 

Holistic approaches regress the 3D translation and orientation components of the target object directly from the given input data~\cite{kehl2017ssd,shi2021stablepose}. 
Xiang et al.~\cite{YuXiang2017PoseCNNAC} introduced a novel PoseCnn framework, which estimated the 3D translation by combining the center location of the target object in the image with the camera parameters. The 3D rotation is regressed from the neural network directly in a quaternion form by utilizing the position and the semantic label of the object.
Li et al.~\cite{li2018deepim} presented a new scheme to regress the 6D pose in an iterative manner, which iteratively refine the pose from initial pose parameters by matching the test image against the rendered image. 
Sundermeyer et al.~\cite{sundermeyer2018implicit} proposed a real-time method to learn an implicit representation of the target, and leveraged an augmented auto-encoder to regress the 6D pose from the latent space. 
Wang et al.~\cite{ChenWang2019DenseFusion6O} designed a novel multi-modal feature fusion framework, which utilizes two feature embedding branches to respectively extract the color and the geometry cues, and employs a pixel-wise feature fusion mechanism to achieve deep interaction of the multi-modality data.
This line of work can achieve satisfactory performance in several benchmarks and be capable to implement end-to-end network architecture directly. However, the generalization and learnability of these methods are restricted by the non-linearity of rotation space.

Current semi-holistic approaches first extract the keypoints on the surface of the target model, and then utilize a PnP algorithm or Least-Squares Fitting~\cite{gupta2019cullnet,rad2017bb8,tekin2018real}. 
Hu et al.~\cite{hu2019segmentation} presented a two-stream segmentation-driven framework, which utilizes each visible cell assigned by the segmentation stream to predict the 2D keypoints locations of the corresponding object. 
Peng et al.~\cite{peng2020pvnet} proposed a novel pixel-wise voting method to predict 2D keypoints of the target object in a given image. The dense keypoints locations are predicted by regressing pixel-wise vectors pointing to the keypoints.
He et al.~\cite{he2021ffb6d} introduced a new bidirectional fusion network, which achieves RGB-D feature fusion effectively and employs the keypoints regression network branch proposed by~\cite{he2020pvn3d} to obtain 3D keyponts.
Keypoint-based methods perform well in occlusion scenarios. However, the detection performance will be severely degraded for texture-less objects when only the RGB images can be used.

\subsection{Equivariant and Invariant Representation Learning}
Recently, equivariance and invariance, as essential properties of point cloud processing in the 3D vision field, have received extensive attention. 

Lin et al.~\cite{lin2021sparse} presented a novel convolution SS-Conv for efficient learning of SE(3)-equivariant features in the point cloud, which designs a sparse steerable kernel based on spherical harmonics function.
Li et al.~\cite{li2021leveraging} proposed a new SE(3)-equivariant point cloud network for category-level object pose estimation, which decouples the pose and object geometry shape by deploying an SE(3)-invariant shape reconstruction module and an SE(3)-equivariant pose estimation module. 
Boulevard et al.~\cite{poulenard2019effective} introduced a novel SO(3)-invariance architecture to process point cloud directly, which employs a spherical harmonics-based kernel at different layers of the network to inject invariant features at the global and local levels. 
Deng et al.~\cite{deng2021vector} introduced a general framework for SO(3)-equivariant feature extraction in the point cloud, which design a vector neurons representation in feature space.

\section{Method} 
Our goal is to estimate the 6D pose parameters of a set of known objects from an RGB-D image. The 6D pose parameters can be represented by a rigid transformation $T\in SE(3)$ from the object coordinate system to the camera coordinate system, which consists of a 3D rotation matrix $R\in SO(3)$ and a 3D translation vector $t\in \mathbb{R}^3$.
Given a depth image $D \in \mathbb{R}^{W\times H}$ where $W$ and $H$ denote the image width and height, and the camera intrinsic matrix $K\in \mathbb{R}^{3 \times 3}$, we can acquire the corresponding point cloud $P\in \mathbb{R}^{(W\times H)\times 3}$ by back-projecting homogeneous pixels $I\in \mathbb{R}^{(W\times H)\times 3} $ in 2D field to 3D space:
\begin{equation}
P = D(x,y)I(K^{-1})^T.
\end{equation}


\subsection{Overview} 
At the top level, we show our SO(3)-Pose in Fig. \ref{fig:overview}. 
SO(3)-Pose follows the high-performing keypoint-based two-stage strategy~\cite{he2020pvn3d,he2021ffb6d}.
A 2D encoder and an SO(3)-equivariant encoder are utilized for representation learning of the
RGB image and point cloud (converted from the depth image), respectively. For better appearance and geometry representation learning, SO(3)-Pose not only absorbs the geometry knowledge of SO(3)-equivariance from the point cloud, but also realizes the cross-modal information communication. By the communication, the SO(3)-invariant features facilitate to learn more distinctive representations for segmenting objects with similar appearance from RGB channels; the SO(3)-equivariant features bridge RGB features to deduce the (missed) geometry for detecting keypoints of an object with the reflective surface from the depth channel. Finally, it utilizes the 3D-3D correspondence of keypoints to fit the 6D pose parameters.

\subsection{SO(3)-equivariant Layer}
\label{equi}
SO(3) can be interpreted as a $3\times3$ rotation matrix.
A feature  $V$ is SO(3)-equivariant if and only if it satisfies:
\begin{equation}
    \PHR{f(VR) = f(V)R,}
\label{so3equa}
\end{equation} 
where $R\in SO(3)$ is a rotation matrix, ${\Large f}$ is a mapping function to represent a layer operation, and $V=\{v _{i} \in \mathbb{R}^{C \times 3}, i=1,2,3,...\}$ ($C$ is the feature dimension). Inspired by VNN \cite{deng2021vector}, our method learns the SO(3)-equivariance properties as follows. 

Learning the rotation equivariance should build a series of basic SO(3)-equivariant layers containing the linear layers, non-linear layers, pooling layers, and normalization layers. These layers all satisfy the SO(3)-equivariance property according to the definition.

Given a weight matrix $\mathbf{W} \in \mathbb{R}^{C^{\prime} \times C}$, we define a linear operation $f_{\operatorname{lin}}(\cdot ; \mathbf{W})$ acting on a vector-list features $\mathbf{V} \in \mathbb{R}^{N\times C \times 3}$ as:
\begin{equation}
\boldsymbol{V}^{\prime}=f_{\operatorname{lin}}(\boldsymbol{V} ; \mathbf{W})=\mathbf{W} \boldsymbol{V} \in \mathbb{R}^{C^{\prime} \times 3}.
\end{equation}
We verify that if the input rotates by $R\in SO(3)$, the output also rotates by the same matrix:
\begin{equation}
\PHR{
f_{\operatorname{lin}}(\boldsymbol{V}R ; \mathbf{W})= \mathbf{W} \boldsymbol{V}R = f_{\operatorname{lin}}(\boldsymbol{V} ; \mathbf{W}) R=  \boldsymbol{V}^{\prime}R,
}
\end{equation}
yielding the desired equivariance property. 

SO(3) has the closure property of a group. That means,  the product of a rotation matrix multiplied by any rotation matrix is also a rotation matrix.
If the final output is SO(3)-equivariant with the input across a neural network, any intermediate output also needs to satisfy the SO(3)-equivariance with the input. Thus, it is necessary to construct a special SO(3)-equivariance layer to achieve the same functions as the non-linear layers, pooling layers and normalization layers. 

Non-linearity plays an important role in the representation learning of neural networks. The non-linear layers (e.g., ReLU, leaky-ReLU) split a feature space into two half-spaces: the positive half-space keeps its original feature and the negative half-space is muted or reduced by multiplication with a small weight. 
To keep SO(3)-equivariance, we dynamically predict a direction from the input vector-list feature and then truncate the portion of a vector that points into the negative half-space of the learned direction:
\begin{equation}
\boldsymbol{v}^{\prime}=\left\{\begin{array}{ll}
\boldsymbol{q} & \text { if }\langle\boldsymbol{q}, \boldsymbol{k}\rangle \geqslant 0 \\
\boldsymbol{q}-\left\langle\boldsymbol{q}, \frac{\boldsymbol{k}}{\|\boldsymbol{k}\|}\right\rangle \frac{\boldsymbol{k}}{\|\boldsymbol{k}\|} & \text { otherwise }
\end{array}\right.
\end{equation}
where $\boldsymbol{q}=\mathbf{W} \boldsymbol{V}$, $\boldsymbol{k}=\mathbf{U} \boldsymbol{V}$, $\mathbf{W} \in \mathbb{R}^{ 1 \times C}$, $\mathbf{U} \in \mathbb{R}^{1 \times C}$ and $\boldsymbol{V} \in \mathbb{R}^{C \times 3}$.

The pooling layer aims to downsample the useful feature.  It also acts as a symmetrical function in the point cloud networks to solve the permutation problem of the point cloud. There are usually two main solutions (i.e., max-pooling and mean-pooling) used in most of the neural networks. 
According to the definition (Eq. \ref{so3equa}), it is satisfied with the average pooling operation. 
We adopt the mean pooling to aggregate all feature information.

The normalization layer influences the convergence efficiency in the training process. Layer normalization \cite{JimmyBa2016LayerN} and instance normalization \cite{DmitryUlyanov2016InstanceNT} only change the distributions in a sample of a batch with respect to a rotation. Batch normalization \cite{SergeyIoffe2015BatchNA} aggregates statistics across all batch samples with respect to several rotations. Averaging across arbitrarily rotated inputs would not necessarily be useful. For example, averaging two input features rotated in opposite directions would be zero instead of producing that feature into a canonical pose. Thus, the operation of batch normalization is required to be specially designed, which cannot break the SO(3)-equivariance of features. We follow the design of VN-Batchnorm \cite{deng2021vector} that is formulated as:
\begin{equation}
    \boldsymbol{N}_{b}=\text { ElementwiseNorm }\left(\boldsymbol{V}_{b}\right) \in \mathbb{R}^{N \times 1},
\end{equation}
\begin{equation}
    \left\{\boldsymbol{N}_{b}^{\prime}\right\}_{b=1}^{B}=\text { BatchNorm }\left(\left\{\boldsymbol{N}_{b}\right\}_{b=1}^{B}\right),
\end{equation}
\begin{equation}
    \boldsymbol{V}_{b}^{\prime}[c]=\boldsymbol{V}_{b}[c] \frac{\boldsymbol{N}_{b}^{\prime}[c]}{\boldsymbol{N}_{b}[c]}, \quad \forall c \in[C],
\end{equation}
where $\left\{\boldsymbol{V}_{b}\right\}_{b=1}^{B}$ are a batch of $B$ vector-list features $\boldsymbol{V}_{b}\in\mathbb{R}^{C \times 3}$, $\boldsymbol{V}_{b}^{\prime}[c]$, $\boldsymbol{V}_{b}[c]$ are the vector channels, $\boldsymbol{V}_{b}^{\prime}[c]$, $\boldsymbol{V}_{b}[c]$ are their scalar 2-norms, and $\text { ElementwiseNorm }\left(\boldsymbol{V}_{b}\right)$ computes the 2-norm of every vector channel  $\boldsymbol{v}_{c}=\boldsymbol{V}_{b}[c] \in \boldsymbol{V}_{b}.$


\begin{figure}[t]

    \centering
    \includegraphics[height=0.5\linewidth]{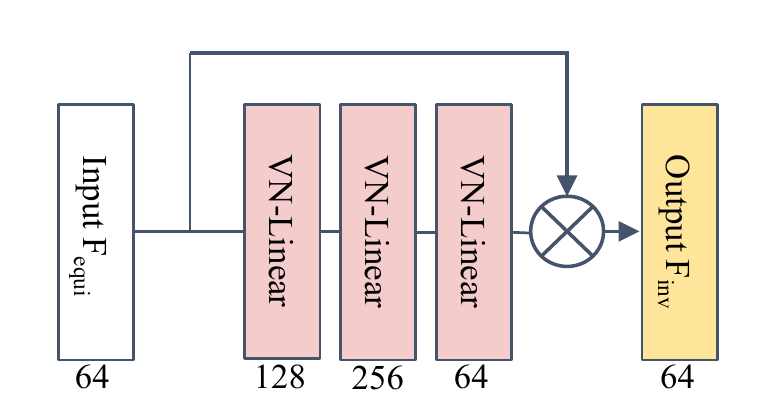}
    \caption{The structure of E2Ilayer. Red box indicates the SO(3)-equivariant linear layer introduced in Section \ref{equi}. }
    \label{fig:E2Ilayer}
\end{figure}
\subsection{SO(3)-invariant Layer}
SO(3)-invariant property is crucial for many tasks such as classification and segmentation, where the label of an object or its parts should be invariant to the object's pose.
We define an SO(3)-equivariant feature as: 
\begin{equation}
     f(VR) = f(V).
\end{equation}
The rotation matrix is an orthonormal matrix; its inverse matrix is equal to its transpose matrix. 
We can compute an SO(3)-invariant feature from two SO(3)-equivariant features:
\begin{equation}
    (\boldsymbol{V} R)(\boldsymbol{T}R)^{\top}=\boldsymbol{V} R R^{\top} \boldsymbol{T}^{\top} =\boldsymbol{V} \boldsymbol{T}^{\top}.
\end{equation}
We propose an E2ILayer (an Equivariant-to-Invariant layer) to learn the SO(3)-invariant feature as illustrated in Fig. \ref{fig:E2Ilayer}. 
We attain the SO(3)-invariant features to map the feature space onto the semantic label space. 
We design a series of linear layers to increase the feature expression ability.

\label{inva}
\subsection{Feature Fusion}
Feature fusion is used to map the SO(3)-equivariant features to the  keypoint offset space and map the SO(3)-invariant features to the semantic label space.
From our observation, the semantic label is invariant when the object rotates, and the relative orientation is changed along with the rotation. From this discovery, we extract the SO(3)-invariant and SO(3)-equivariant features from the point cloud. 
We use a general 2D extractor from RGB images to contain local and global appearance features, then we fuse the appearance and SO(3)-equivariant features to map the feature space onto the keypoint offset space. 
We fuse the appearance features and SO(3)-invariant features to map the feature space onto the semantic space. Specifically, we first concatenate the geometric feature and appearance feature, and feed the fused features to MLP to generate the mixed feature space for final prediction. 
To avoid overfitting, we only use two linear layers. 

\label{fusi}
\subsection{6D Object Pose Estimation}
Once each point is assigned with a semantic label and an offset relative to the keypoints, we can start the second stage of pose estimation. 

In detail, we first attain the object instance according to the per-point label and center point which is one of the predicted keypoints. 
Next, we cluster the keypoints to vote for certain keypoints corresponding to the keypoints selected from the object models. Finally, we utilize a least-squares fitting algorithm based on the 3D-3D corresponding relation to compute the pose parameters. 


Let $L_{\text {seg}}$ and $L_{\text {kp}}$ be the two losses that are imposed on the segmentation branch and keypoint detection branch. 
For the special center point, we additionally use $L_{\text {center}}$. $L_{\text {seg}}$ is a focal loss \cite{TsungYiLin2017FocalLF}. 
$L_{\text {kp}}$ and $L_{\text {center}}$ are $L_1$ distance losses following PVN3D \cite{he2020pvn3d}. 
Features extracted from the SO(3)-equivariant layers have the transitive dependencies property, 
in that each former feature rotates by $R$ is equivalent to all latter features rotated by the same matrix $R$. 
In order to improve the expression of the SO(3)-equivariant property, we propose a new loss function, dubbed SO3Loss, for the SO(3)-equivariant layers as:
\begin{equation}
    L_{\text {so3}}= \left|f(V)-f(VR)R^{\mbox{-}1}\right|,
\end{equation}
where $f$ represents a mapping through a series of SO(3)-equivariant layers, $R$ is any rotation matrix, and $V$ is a former feature.  We jointly optimize the detection task and the auxiliary tasks by applying a gradient descent method to minimize the weighted sum of the following losses:
\begin{equation}
    L_{\text {all }}=\lambda_{1} L_{\text {seg }}+\lambda_{2} L_{\text {kp }}+\lambda_{3} L_{\text {center }}+\lambda_{4} L_{\text {so3 }},
\end{equation}
where $\lambda_{1}$, $\lambda_{2}$, $\lambda_{3}$ and $\lambda_{4}$ are the weights for each task. 

\label{so3loss}

\section{Experiments}
\subsection{DataSet}
We evaluate our method on two benchmark datasets, including the YCB-Video dataset and the LineMOD dataset. 

\textbf{YCB-Video dataset.} YCB-Video \cite{YuXiang2017PoseCNNAC} consists of 21 YCB objects \cite{BerkCalli2015TheYO} in 92 RGB-D videos. All the subsets of the objects that appeared in the scene are annotated with 6D pose and instance-level masks. We follow the previous works  \cite{YuXiang2017PoseCNNAC, ChenWang2019DenseFusion6O, he2020pvn3d} to split the training set and the testing set. 
We also take the synthetic images for training as per \cite{YuXiang2017PoseCNNAC} and apply the hole completion algorithm to fill the depth images as per \cite{JasonKu2018InDO}. 

\textbf{LineMOD dataset.} LineMOD \cite{StefanHinterstoisser2011MultimodalTF} contains 13 low-textured objects in 13 videos, with annotated 6D pose and instance-semantic masks. The varying lighting, texture-less objects and cluttered scenes make this dataset challenging. We follow prior works \cite{YuXiang2017PoseCNNAC, ChenWang2019DenseFusion6O} to split the training set and the testing set, and we also obtain synthesis images for the training set following \cite{peng2019pvnet, he2020pvn3d}.

\textbf{Occlusion LineMOD dataset.} Occlusion LineMOD \cite{EricBrachmann2014Learning6O} is created by additionally annotating a subset of the LineMOD dataset. It contains 8 objects which are out of 13 LineMOD objects. The objects in heavily occlusion makes the dataset challenging.

\begin{table}[t]
\centering

\begin{tabular}{l|ccc}
\hline
\multicolumn{1}{c|}{method} & E              & E+F            & E+F+S          \\ \hline
002\_master\_chef\_can      & 95.35          & 95.10          & \textbf{95.52} \\
003\_cracker\_box             & \textbf{95.19} & 94.86          & 94.32          \\
004\_sugar\_box               & 96.53          & \textbf{96.60} & 96.47          \\
005\_tomato\_soup\_can         & 95.19          & 95.15          & \textbf{95.35} \\
006\_mustard\_bottle          & 95.94          & 96.15          & \textbf{96.74} \\
007\_tuna\_fish\_can           & 95.64          & \textbf{96.05} & 95.80          \\
008\_pudding\_box             & 96.47          & 96.18          & \textbf{96.83} \\
009\_gelatin\_box             & 97.00          & \textbf{97.54} & 97.30          \\
010\_potted\_meat\_can         & 93.45          & \textbf{93.47} & 92.42          \\
011\_banana                  & 94.61          & 95.05          & \textbf{95.75} \\
019\_pitcher\_base            & 95.55          & \textbf{95.95} & 95.56          \\
021\_bleach\_cleanser         & \textbf{95.41} & 94.90          & 95.14          \\
\textbf{024\_bowl}                    & 81.01          & 84.78          & \textbf{88.41} \\
025\_mug                     & 96.88          & \textbf{96.92} & 96.72          \\
035\_power\_drill             & 95.42          & 95.84          & \textbf{95.90} \\
\textbf{036\_wood\_block}              & \textbf{87.54} & 83.12          & 86.58          \\
037\_scissors                & 90.90          & \textbf{95.43} & 91.84          \\
040\_large\_marker            & 93.76          & \textbf{95.00} & 94.37          \\
\textbf{051\_large\_clamp}             & 88.48          & 91.93          & \textbf{94.15} \\
\textbf{052\_extra\_large\_clamp}       & 84.51          & 90.99          & \textbf{91.22} \\
\textbf{061\_foam\_brick}              & 93.94          & \textbf{95.30} & 94.98          \\ \hline
average                     & 93.57          & 94.11          & \textbf{94.35}
\end{tabular}
\caption{ Ablation studies on different configurations in terms of the ADDS metric, where objects in bold are considered as the symmetric objects.  ``E'' only uses the SO(3)-equivariant layers. ``E+F'' adds the SO(3)-invariat layers and the feature fusion strategy, ``E+F+S'' further introduces the SO3loss. }
\label{ycb_abl}
\end{table}
\begin{table*}[h]
\centering

\resizebox{\textwidth}{!}{

\begin{tabular}{l|cc|cc|cc|cc|cc|cc|cc}
\hline
\multicolumn{1}{c|}{}             & \multicolumn{2}{c|}{PoseCNN}        & \multicolumn{2}{c|}{PoseCNN+ICP}                   & \multicolumn{2}{c|}{DF(per-pixel)}  & \multicolumn{2}{c|}{DF(iterative)}                 & \multicolumn{2}{c|}{FFB6D(trained)}                & \multicolumn{2}{c|}{Our}                  & \multicolumn{2}{c}{Our+ICP}                        \\ \hline
                                  & \multicolumn{1}{c|}{ADDS} & ADD(-S) & \multicolumn{1}{c|}{ADDS}          & ADD(-S)       & \multicolumn{1}{c|}{ADDS} & ADD(-S) & \multicolumn{1}{c|}{ADDS}          & ADD(-S)       & \multicolumn{1}{c|}{ADDS}          & ADD(-S)       & \multicolumn{1}{c|}{ADDS} & ADD(-S)       & \multicolumn{1}{c|}{ADDS}          & ADD(-S)       \\ \hline
002\_master\_chef\_can            & \multicolumn{1}{c|}{83.9} & 50.2    & \multicolumn{1}{c|}{95.8}          & 68.1          & \multicolumn{1}{c|}{95.3} & 70.7    & \multicolumn{1}{c|}{\textbf{96.4}} & 73.2          & \multicolumn{1}{c|}{95.7}          & 78.6          & \multicolumn{1}{c|}{95.5} & 82.2          & \multicolumn{1}{c|}{95.5}          & \textbf{82.8} \\
003\_cracker\_box                 & \multicolumn{1}{c|}{76.9} & 53.1    & \multicolumn{1}{c|}{92.7}          & 83.4          & \multicolumn{1}{c|}{92.5} & 86.9    & \multicolumn{1}{c|}{\textbf{95.8}} & \textbf{94.1} & \multicolumn{1}{c|}{94.4}          & 90.0          & \multicolumn{1}{c|}{94.3} & 90.6          & \multicolumn{1}{c|}{91.1}          & 85.1          \\
004\_sugar\_box                   & \multicolumn{1}{c|}{84.2} & 68.4    & \multicolumn{1}{c|}{\textbf{98.2}} & \textbf{97.1} & \multicolumn{1}{c|}{95.1} & 90.8    & \multicolumn{1}{c|}{95.8}          & 94.1          & \multicolumn{1}{c|}{96.3}          & 93.4          & \multicolumn{1}{c|}{96.5} & 94.2          & \multicolumn{1}{c|}{97.3}          & 96.3          \\
005\_tomato\_soup\_can            & \multicolumn{1}{c|}{81.0} & 66.2    & \multicolumn{1}{c|}{94.5}          & 81.8          & \multicolumn{1}{c|}{93.8} & 84.7    & \multicolumn{1}{c|}{94.5}          & 85.5          & \multicolumn{1}{c|}{94.3}          & 83.1          & \multicolumn{1}{c|}{95.4} & 89.5          & \multicolumn{1}{c|}{\textbf{95.7}} & \textbf{89.9} \\
006\_mustard\_bottle              & \multicolumn{1}{c|}{90.4} & 81.0    & \multicolumn{1}{c|}{\textbf{98.6}} & \textbf{98.0} & \multicolumn{1}{c|}{95.8} & 90.9    & \multicolumn{1}{c|}{97.3}          & 94.7          & \multicolumn{1}{c|}{96.7}          & 94.4          & \multicolumn{1}{c|}{96.7} & 94.6          & \multicolumn{1}{c|}{97.8}          & 97.0          \\
007\_tuna\_fish\_can              & \multicolumn{1}{c|}{88.0} & 70.7    & \multicolumn{1}{c|}{97.1}          & 83.9          & \multicolumn{1}{c|}{95.7} & 79.6    & \multicolumn{1}{c|}{97.1}          & 81.9          & \multicolumn{1}{c|}{95.5}          & 83.1          & \multicolumn{1}{c|}{95.8} & 86.6          & \multicolumn{1}{c|}{\textbf{97.1}} & \textbf{87.8} \\
008\_pudding\_box                 & \multicolumn{1}{c|}{79.1} & 62.7    & \multicolumn{1}{c|}{\textbf{97.9}} & \textbf{96.6} & \multicolumn{1}{c|}{94.3} & 89.3    & \multicolumn{1}{c|}{96.0}          & 93.3          & \multicolumn{1}{c|}{95.5}          & 91.6          & \multicolumn{1}{c|}{96.8} & 94.2          & \multicolumn{1}{c|}{97.5}          & 96.1          \\
009\_gelatin\_box                 & \multicolumn{1}{c|}{87.2} & 75.2    & \multicolumn{1}{c|}{\textbf{98.8}} & \textbf{98.1} & \multicolumn{1}{c|}{97.2} & 95.8    & \multicolumn{1}{c|}{98.0}          & 96.7          & \multicolumn{1}{c|}{96.8}          & 93.5          & \multicolumn{1}{c|}{97.3} & 94.7          & \multicolumn{1}{c|}{98.5}          & 97.8          \\
010\_potted\_meat\_can            & \multicolumn{1}{c|}{78.5} & 59.5    & \multicolumn{1}{c|}{\textbf{92.7}} & 83.5          & \multicolumn{1}{c|}{89.3} & 79.6    & \multicolumn{1}{c|}{90.7}          & 83.6          & \multicolumn{1}{c|}{89.7}          & 82.9          & \multicolumn{1}{c|}{92.4} & \textbf{84.5} & \multicolumn{1}{c|}{91.6}          & 80.9          \\
011\_banana                       & \multicolumn{1}{c|}{86.0} & 72.3    & \multicolumn{1}{c|}{97.1}          & 91.9          & \multicolumn{1}{c|}{90.0} & 76.7    & \multicolumn{1}{c|}{96.2}          & 83.3          & \multicolumn{1}{c|}{96.8}          & 94.0          & \multicolumn{1}{c|}{95.8} & 91.3          & \multicolumn{1}{c|}{\textbf{97.5}} & \textbf{95.1} \\
019\_pitcher\_base                & \multicolumn{1}{c|}{77.0} & 53.3    & \multicolumn{1}{c|}{\textbf{97.8}} & 96.9          & \multicolumn{1}{c|}{93.6} & 87.1    & \multicolumn{1}{c|}{97.5}          & \textbf{96.9} & \multicolumn{1}{c|}{95.5}          & 91.4          & \multicolumn{1}{c|}{95.6} & 91.7          & \multicolumn{1}{c|}{97.1}          & 95.8          \\
021\_bleach\_cleanser             & \multicolumn{1}{c|}{71.6} & 50.3    & \multicolumn{1}{c|}{96.9}          & 92.5          & \multicolumn{1}{c|}{94.4} & 87.5    & \multicolumn{1}{c|}{95.9}          & 89.9          & \multicolumn{1}{c|}{95.4}          & 90.6          & \multicolumn{1}{c|}{95.1} & 91.4          & \multicolumn{1}{c|}{\textbf{96.9}} & \textbf{94.9} \\
\textbf{024\_bowl}                & \multicolumn{1}{c|}{69.6} & 69.6    & \multicolumn{1}{c|}{81.0}          & 81.0          & \multicolumn{1}{c|}{86.0} & 86.0    & \multicolumn{1}{c|}{\textbf{89.5}} & \textbf{89.5} & \multicolumn{1}{c|}{86.2}          & 86.2          & \multicolumn{1}{c|}{88.4} & 88.4          & \multicolumn{1}{c|}{87.8}          & 87.8          \\
025\_mug                          & \multicolumn{1}{c|}{78.2} & 58.5    & \multicolumn{1}{c|}{94.9}          & 81.1          & \multicolumn{1}{c|}{95.3} & 83.8    & \multicolumn{1}{c|}{96.7}          & 88.9          & \multicolumn{1}{c|}{97.0}          & 91.3          & \multicolumn{1}{c|}{96.7} & 91.1          & \multicolumn{1}{c|}{\textbf{97.7}} & \textbf{94.8} \\
035\_power\_drill                 & \multicolumn{1}{c|}{72.7} & 55.3    & \multicolumn{1}{c|}{\textbf{98.2}} & \textbf{97.7} & \multicolumn{1}{c|}{92.1} & 83.7    & \multicolumn{1}{c|}{96.0}          & 92.7          & \multicolumn{1}{c|}{95.9}          & 93.1          & \multicolumn{1}{c|}{95.9} & 93.2          & \multicolumn{1}{c|}{96.6}          & 94.8          \\
\textbf{036\_wood\_block}         & \multicolumn{1}{c|}{64.3} & 64.3    & \multicolumn{1}{c|}{87.6}          & 87.6          & \multicolumn{1}{c|}{89.5} & 89.5    & \multicolumn{1}{c|}{\textbf{92.8}} & \textbf{92.8} & \multicolumn{1}{c|}{91.5}          & 91.5          & \multicolumn{1}{c|}{86.6} & 86.6          & \multicolumn{1}{c|}{91.6}          & 91.6          \\
037\_scissors                     & \multicolumn{1}{c|}{56.9} & 35.8    & \multicolumn{1}{c|}{91.7}          & 78.4          & \multicolumn{1}{c|}{90.1} & 77.4    & \multicolumn{1}{c|}{92.0}          & 77.9          & \multicolumn{1}{c|}{\textbf{94.1}} & \textbf{83.0} & \multicolumn{1}{c|}{91.8} & 81.0          & \multicolumn{1}{c|}{88.5}          & 73.1          \\
040\_large\_marker                & \multicolumn{1}{c|}{71.7} & 58.3    & \multicolumn{1}{c|}{97.2}          & 85.3          & \multicolumn{1}{c|}{95.1} & 89.1    & \multicolumn{1}{c|}{97.6}          & 93.0          & \multicolumn{1}{c|}{94.7}          & 85.2          & \multicolumn{1}{c|}{94.4} & 86.9          & \multicolumn{1}{c|}{\textbf{97.6}} & \textbf{88.3} \\
\textbf{051\_large\_clamp}        & \multicolumn{1}{c|}{50.2} & 50.2    & \multicolumn{1}{c|}{75.2}          & 75.2          & \multicolumn{1}{c|}{71.5} & 71.5    & \multicolumn{1}{c|}{72.5}          & 72.5          & \multicolumn{1}{c|}{91.0}          & 91.0          & \multicolumn{1}{c|}{94.2} & 94.2          & \multicolumn{1}{c|}{\textbf{95.8}} & \textbf{95.8} \\
\textbf{052\_extra\_large\_clamp} & \multicolumn{1}{c|}{44.1} & 44.1    & \multicolumn{1}{c|}{64.4}          & 64.4          & \multicolumn{1}{c|}{70.2} & 70.2    & \multicolumn{1}{c|}{69.9}          & 69.9          & \multicolumn{1}{c|}{91.2}          & 91.2          & \multicolumn{1}{c|}{91.2} & 91.2          & \multicolumn{1}{c|}{\textbf{91.7}} & \textbf{91.7} \\
\textbf{061\_foam\_brick}         & \multicolumn{1}{c|}{88.0} & 88.0    & \multicolumn{1}{c|}{\textbf{97.2}} & \textbf{97.2} & \multicolumn{1}{c|}{92.2} & 92.2    & \multicolumn{1}{c|}{92.0}          & 92.0          & \multicolumn{1}{c|}{93.8}          & 93.8          & \multicolumn{1}{c|}{95.0} & 95.0          & \multicolumn{1}{c|}{97.0}          & 97.0          \\ \hline
average                           & \multicolumn{1}{c|}{75.8} & 59.9    & \multicolumn{1}{c|}{93.0}          & 85.4          & \multicolumn{1}{c|}{91.2} & 82.9    & \multicolumn{1}{c|}{93.2}          & 86.1          & \multicolumn{1}{c|}{94.2}          & 89.1          & \multicolumn{1}{c|}{94.4} & 90.1          & \multicolumn{1}{c|}{\textbf{95.1}} & \textbf{91.1} \\ \hline
\end{tabular}
}
\caption{Quantitative evaluation of ADDS AUC and ADD(-S) AUC metrics on the YCB-Video dataset. Objects in bold are considered as the symmetric objects. FFB6D(trained) indicates the model is trained 
under its original settings \cite{he2021ffb6d}. }
\label{ycb_other}
\end{table*}
\begin{table*}[ht]
\centering
\resizebox{\textwidth}{!}{
\begin{tabular}{l|cccc|ccccccc}
\hline
\multicolumn{1}{c|}{} & \multicolumn{4}{c|}{RGB}                                                                                                                                 & \multicolumn{7}{c}{RGB-D}                                                                                                                                                                                                                                                                                                     \\ \hline
                      & \multicolumn{1}{c|}{\begin{tabular}[c]{@{}c@{}}PoseCNN+\\ DeepIM\end{tabular}} & \multicolumn{1}{c|}{PVNet} & \multicolumn{1}{c|}{CDPN} & DPOD           & \multicolumn{1}{c|}{\begin{tabular}[c]{@{}c@{}}Point-\\ Fusion\end{tabular}} & \multicolumn{1}{c|}{\begin{tabular}[c]{@{}c@{}}Dense-\\ Fusion\end{tabular}} & \multicolumn{1}{c|}{G2LNet}         & \multicolumn{1}{c|}{PVN3D}          & \multicolumn{1}{c|}{SS-Conv} & \multicolumn{1}{c|}{FFB6D}          & our            \\ \hline
ape                   & \multicolumn{1}{c|}{77.0}                                                      & \multicolumn{1}{c|}{43.6}  & \multicolumn{1}{c|}{64.4} & 87.7           & \multicolumn{1}{c|}{70.4}                                                    & \multicolumn{1}{c|}{92.3}                                                    & \multicolumn{1}{c|}{96.8}           & \multicolumn{1}{c|}{97.3}           & \multicolumn{1}{c|}{97.4}    & \multicolumn{1}{c|}{98.4}           & \textbf{98.6}  \\
benchvise             & \multicolumn{1}{c|}{97.5}                                                      & \multicolumn{1}{c|}{99.9}  & \multicolumn{1}{c|}{97.8} & 98.5           & \multicolumn{1}{c|}{80.7}                                                    & \multicolumn{1}{c|}{93.2}                                                    & \multicolumn{1}{c|}{96.1}           & \multicolumn{1}{c|}{99.7}           & \multicolumn{1}{c|}{99.3}    & \multicolumn{1}{c|}{\textbf{100.0}} & \textbf{100.0} \\
camera                & \multicolumn{1}{c|}{93.5}                                                      & \multicolumn{1}{c|}{86.9}  & \multicolumn{1}{c|}{91.7} & 96.1           & \multicolumn{1}{c|}{60.8}                                                    & \multicolumn{1}{c|}{94.4}                                                    & \multicolumn{1}{c|}{98.2}           & \multicolumn{1}{c|}{99.6}           & \multicolumn{1}{c|}{99.5}    & \multicolumn{1}{c|}{\textbf{99.9}}  & \textbf{99.9}  \\
can                   & \multicolumn{1}{c|}{96.5}                                                      & \multicolumn{1}{c|}{95.5}  & \multicolumn{1}{c|}{95.9} & 99.7           & \multicolumn{1}{c|}{61.1}                                                    & \multicolumn{1}{c|}{93.1}                                                    & \multicolumn{1}{c|}{98.0}           & \multicolumn{1}{c|}{99.5}           & \multicolumn{1}{c|}{99.6}    & \multicolumn{1}{c|}{99.8}           & \textbf{100.0} \\
cat                   & \multicolumn{1}{c|}{82.1}                                                      & \multicolumn{1}{c|}{79.3}  & \multicolumn{1}{c|}{83.8} & 94.7           & \multicolumn{1}{c|}{79.1}                                                    & \multicolumn{1}{c|}{96.5}                                                    & \multicolumn{1}{c|}{99.2}           & \multicolumn{1}{c|}{99.8}           & \multicolumn{1}{c|}{99.8}    & \multicolumn{1}{c|}{99.9}           & \textbf{100.0} \\
driller               & \multicolumn{1}{c|}{95.0}                                                      & \multicolumn{1}{c|}{96.4}  & \multicolumn{1}{c|}{96.2} & 98.8           & \multicolumn{1}{c|}{47.3}                                                    & \multicolumn{1}{c|}{87.0}                                                    & \multicolumn{1}{c|}{99.8}           & \multicolumn{1}{c|}{99.3}           & \multicolumn{1}{c|}{99.6}    & \multicolumn{1}{c|}{100.0}          & \textbf{100.0} \\
duck                  & \multicolumn{1}{c|}{77.7}                                                      & \multicolumn{1}{c|}{52.6}  & \multicolumn{1}{c|}{66.8} & 86.3           & \multicolumn{1}{c|}{63.0}                                                    & \multicolumn{1}{c|}{92.3}                                                    & \multicolumn{1}{c|}{97.7}           & \multicolumn{1}{c|}{98.2}           & \multicolumn{1}{c|}{97.8}    & \multicolumn{1}{c|}{98.4}           & \textbf{98.5}  \\
\textbf{eggbox}       & \multicolumn{1}{c|}{97.1}                                                      & \multicolumn{1}{c|}{99.2}  & \multicolumn{1}{c|}{99.7} & 99.9           & \multicolumn{1}{c|}{99.9}                                                    & \multicolumn{1}{c|}{99.8}                                                    & \multicolumn{1}{c|}{\textbf{100.0}} & \multicolumn{1}{c|}{99.8}           & \multicolumn{1}{c|}{99.9}    & \multicolumn{1}{c|}{\textbf{100.0}} & \textbf{100.0} \\
\textbf{glue}         & \multicolumn{1}{c|}{99.4}                                                      & \multicolumn{1}{c|}{95.7}  & \multicolumn{1}{c|}{99.6} & 96.8           & \multicolumn{1}{c|}{99.3}                                                    & \multicolumn{1}{c|}{\textbf{100.0}}                                          & \multicolumn{1}{c|}{\textbf{100.0}} & \multicolumn{1}{c|}{\textbf{100.0}} & \multicolumn{1}{c|}{99.6}    & \multicolumn{1}{c|}{\textbf{100.0}} & \textbf{100.0} \\
holepuncher           & \multicolumn{1}{c|}{52.8}                                                      & \multicolumn{1}{c|}{82.0}  & \multicolumn{1}{c|}{85.8} & 86.9           & \multicolumn{1}{c|}{71.8}                                                    & \multicolumn{1}{c|}{92.1}                                                    & \multicolumn{1}{c|}{99.0}           & \multicolumn{1}{c|}{99.9}           & \multicolumn{1}{c|}{99.4}    & \multicolumn{1}{c|}{99.8}           & \textbf{100.0} \\
iron                  & \multicolumn{1}{c|}{98.3}                                                      & \multicolumn{1}{c|}{98.9}  & \multicolumn{1}{c|}{97.9} & \textbf{100.0} & \multicolumn{1}{c|}{83.2}                                                    & \multicolumn{1}{c|}{97.0}                                                    & \multicolumn{1}{c|}{99.3}           & \multicolumn{1}{c|}{99.7}           & \multicolumn{1}{c|}{99.2}    & \multicolumn{1}{c|}{99.9}           & \textbf{100.0} \\
lamp                  & \multicolumn{1}{c|}{97.5}                                                      & \multicolumn{1}{c|}{99.3}  & \multicolumn{1}{c|}{97.9} & 96.8           & \multicolumn{1}{c|}{62.3}                                                    & \multicolumn{1}{c|}{95.3}                                                    & \multicolumn{1}{c|}{99.5}           & \multicolumn{1}{c|}{99.8}           & \multicolumn{1}{c|}{99.7}    & \multicolumn{1}{c|}{99.9}           & \textbf{100.0} \\
phone                 & \multicolumn{1}{c|}{87.7}                                                      & \multicolumn{1}{c|}{92.4}  & \multicolumn{1}{c|}{90.8} & 94.7           & \multicolumn{1}{c|}{78.8}                                                    & \multicolumn{1}{c|}{92.8}                                                    & \multicolumn{1}{c|}{98.9}           & \multicolumn{1}{c|}{99.5}           & \multicolumn{1}{c|}{98.2}    & \multicolumn{1}{c|}{99.7}           & \textbf{99.9}  \\ \hline
MEAN                  & \multicolumn{1}{c|}{88.6}                                                      & \multicolumn{1}{c|}{86.3}  & \multicolumn{1}{c|}{89.9} & 95.2           & \multicolumn{1}{c|}{73.7}                                                    & \multicolumn{1}{c|}{94.3}                                                    & \multicolumn{1}{c|}{98.7}           & \multicolumn{1}{c|}{99.4}           & \multicolumn{1}{c|}{99.2}    & \multicolumn{1}{c|}{99.7}           & \textbf{99.8}  \\ \hline
\end{tabular}
}
\caption{\PHR{Quantitative evaluation using the ADD(-S)-0.1d metric on the LineMOD dataset. Objects with bold name are symmetric.}}
\label{lm_other}
\end{table*}
\begin{table*}[ht]
\centering
\resizebox{\textwidth}{!}{
\begin{tabular}{l|c|c|c|c|c|c|c|c|c}
\hline
\multicolumn{1}{c|}{Method} & PoseCNN & Pix2Pose & PVNet & DPOD & \begin{tabular}[c]{@{}c@{}}Hu et \\ al.\end{tabular} & HybridPose & PVN3D         & FFB6D         & our           \\ \hline
ape                         & 9.6     & 22.0     & 15.8  & -    & 19.2                                                 & 20.9       & 33.9          & 47.2          & \textbf{49.7} \\
can                         & 45.2    & 44.7     & 63.3  & -    & 65.1                                                 & 75.3       & 88.6          & 85.2          & \textbf{88.8} \\
cat                         & 0.9     & 22.7     & 16.7  & -    & 18.9                                                 & 24.9       & 39.1          & 45.7          & \textbf{50.9} \\
driller                     & 41.4    & 44.7     & 65.7  & -    & 69.0                                                 & 70.2       & 78.4          & 81.4          & \textbf{88.6} \\
duck                        & 19.6    & 15.0     & 25.2  & -    & 25.3                                                 & 27.9       & 41.9          & 53.9          & \textbf{58.1} \\
\textbf{eggbox}             & 22.0    & 25.2     & 50.2  & -    & 52.0                                                 & 52.4       & \textbf{80.9} & 70.2          & 60.8          \\
\textbf{glue}               & 38.5    & 32.4     & 49.6  & -    & 51.4                                                 & 53.8       & 68.1          & 60.1          & \textbf{71.0} \\
holepuncher                 & 22.1    & 49.5     & 39.7  & -    & 45.6                                                 & 54.2       & 74.7          & \textbf{85.9} & 82.4          \\ \hline
MEAN                        & 24.9    & 32.0     & 40.8  & 47.3 & 43.3                                                 & 47.5       & 63.2          & 66.2          & \textbf{68.4} \\ \hline
\end{tabular}
}
\caption{\PHR{Quantitative evaluation using the ADD(-S)-0.1d metric on the Occlusion LineMOD dataset. Hu et al. means the paper \cite{YinlinHu2020SingleStage6O}. Symmetric objects' names are in bold.}}
\label{occlm_other}
\end{table*}

\subsection{Metrics}
We evaluate our method with the average distance metrics ADD \cite{StefanHinterstoisser2011MultimodalTF} and ADD-S \cite{YuXiang2017PoseCNNAC}. For non-symmetric objects, the ADD metric calculates the point-pair average distance between object model vertices transformed by the ground truth pose $[R^{*}, t^{*}]$ and the predicted pose $[R, t]$:
\begin{equation}
    \mathrm{ADD}=\frac{1}{m} \sum_{x \in \mathcal{O}}\left\|(R^{*} x+t^{*})-\left(R x+t\right)\right\|.
\end{equation}
where $x$ denotes a vertex in the object model $\mathcal{O}$, m is the number of vertices. For symmetric objects, the ADD-S metric calculates the mean distance based on the closest point distance: 
\begin{equation}
    \mathrm{ADD}\mbox{-} \mathrm{S}=\frac{1}{m} \sum_{x_{1} \in \mathcal{O}} \min _{x_{2} \in \mathcal{O}}\left\|\left(R^{*} x_{1}+t^{*}\right)-\left(R x_{2}+t\right)\right\|.
\end{equation}

In the YCB-Video dataset, we follow prior works \cite{YuXiang2017PoseCNNAC, ChenWang2019DenseFusion6O, he2021ffb6d} to report the area under the ADDS and ADD(-S) \cite{hinterstoisser2012model} curve (AUC) and set the maximum threshold of AUC to be 0.1m. The ADD(-S) calculates ADD for non-symmetric objects and ADDS for symmetric objects. In the LineMOD dataset, we follow \cite{peng2019pvnet} to use the ADD(-S)-0.1d which indicates that the estimated pose is correct When the ADD(-S) distance is less than 10\% of the model's diameter. 
\subsection{ Implementation Details}
We train our model on the original YCB-Video dataset and the additional rendering LineMOD dataset following FFB6D \cite{he2021ffb6d}. We select the PSPNet \cite{HengshuangZhao2017PyramidSP} as the 2D feature extractor which is widely used in the field of image segmentation. We design the point cloud feature extractor by the component presented by VNN \cite{deng2021vector}. 
In the training stage, we set the loss function weights $\lambda_{1}$, $\lambda_{2}$, $\lambda_{3}$ and $\lambda_{4}$ as 1.0, 1.0, 1.0, 0.5, respectively. We train by 20 epochs in the YCB-Video dataset and 10 epochs in the LineMOD dataset for single objects.

\subsection{Ablation Studies}
We conduct ablation studies to evaluate the effect of the feature fusion strategy and the SO(3) loss on the YCB-Video dataset. Table \ref{ycb_abl} summarizes the results of ablation studies on the YCB-video dataset.

The column ``E'' is based on the SO(3)-equivariance layers and uses the concatenation of SO(3)-equivariant features and appearance features to relate both the semantic label and keypoint offset. Column ``E+F'' denotes the base SO(3)-equivariance layer and an additional feature fusion strategy. 
Column ``E+F+S'' shows the results of further adding the SO(3) loss. The 2D encoder is the same for all models.

To validate the benefit of the feature fusion strategy, we compare the column ``E+F'' with the column ``E''. The results show that the feature fusion strategy can learn a better mapping from the feature space to the semantic space and keypoint offset space, thus increasing the accuracy of pose estimation. 

To analyze the SO(3) loss, we compare the pose estimation results based on the ``E+F'' model. The results in Column ``E+F+S'' demonstrate that adding the SO(3) loss improves the accuracy of pose estimation. 

\begin{figure*}[ht]
    \centering

    \includegraphics[width=1.0\textwidth]{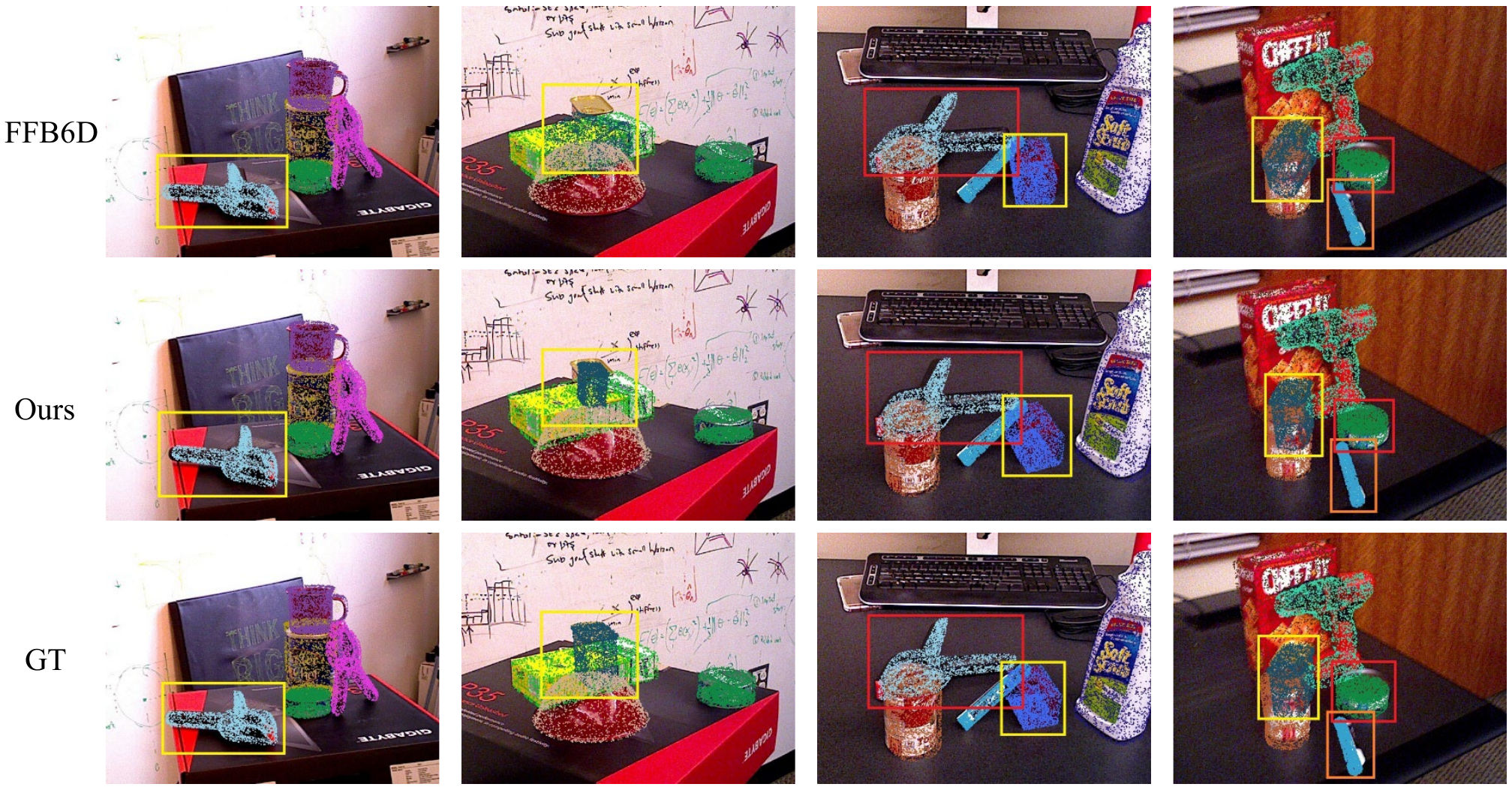}

    \caption{Visual Comparison with FFB6D \cite{he2021ffb6d} on YCB-Video. Different objects in the same scene are in different colors. The points are projected back to the image after being transformed by the predicted pose. 
    }
    \label{vis}
\end{figure*}


\begin{figure*}[ht]
    \centering

    \includegraphics[width=1.0\textwidth]{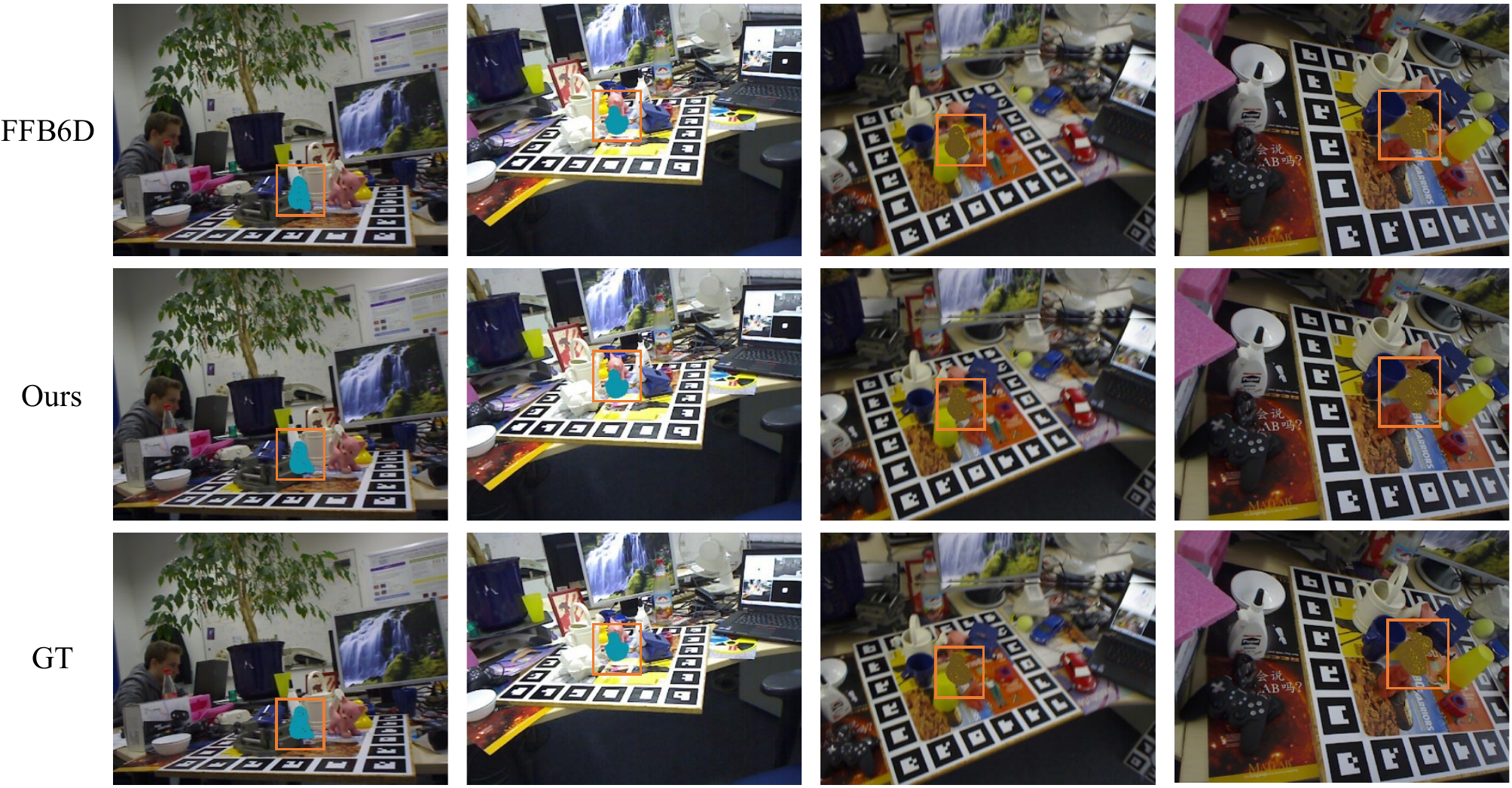}

    \caption{Visual Comparison with FFB6D \cite{he2021ffb6d}. We select two challenging objects (ape boxed in blue and duck boxed in orange) from LineMOD. 
    }
    \label{lmvis}
\end{figure*}
\subsection{Comparisons}
We compare our method with the state-of-the-art methods which take RGB or RGB-D as input and output the 6D object pose. 

\textbf{Performance on the YCB-Video dataset.} In Table \ref{ycb_other}, we compare our method with PoseCNN \cite{YuXiang2017PoseCNNAC}, DenseFusion \cite{ChenWang2019DenseFusion6O}, FFB6D \cite{he2021ffb6d} on the YCB-Video dataset in terms of the ADDS AUC metric and ADD(-S) AUC metric. Our method achieves competitive performance with its competitors. \cite{YuXiang2017PoseCNNAC, ChenWang2019DenseFusion6O} directly regress the rotation matrix and translation vector, while our two-stage method 
first predicts the keypoint localization and then fits the 6D pose parameters. 
The results show that our method performs well for the large clamp and the extra-large clamp which are difficult to detect because of their symmetry and similarity. \PHR{Comparing with other methods with refinement procedure, our model with ICP outperforms PoseCNN+ICP by 6.7\% and exceeds DF(iterative) by 5.8\% on the ADD(-S) metric.}

\textbf{Performance on the LineMOD dataset.} In Table \ref{lm_other}, we compare our method with PoseCNN \cite{YuXiang2017PoseCNNAC} + DeepIM \cite{li2018deepim}, PointFusion \cite{DanfeiXu2017PointFusionDS}, PVNet \cite{peng2019pvnet}, DenseFusion \cite{ChenWang2019DenseFusion6O}, G2LNet \cite{WeiChen2020G2LNetGT}, PVN3D \cite{he2020pvn3d}, FFB6D \cite{he2021ffb6d} and SS-Conv \cite{lin2021sparse} on the LineMOD dataset in terms of the ADD(-S)-0.1d metrics. Our method achieves the state-of-the-art performance. 
\cite{YuXiang2017PoseCNNAC, peng2019pvnet} are based on the RGB images, and the other methods are based on the RGB-D images. 
G2LNet \cite{WeiChen2020G2LNetGT} first learns the global feature, 
and then learns the local feature. SS-Conv \cite{lin2021sparse} learns the SE(3)-equivariant feature to relate the rotation space and the translation space. The results show that our design of learning SO(3)-equivariant features 
outperforms its competitors. On the ADD(-S)-0.1d metric, our model achieves the SOTA at all objects.

\PHR{
\textbf{Performance on the Occlusion LineMOD dataset.} We use the model trained on the LineMOD dataset for testing on the Occlusion LineMOD dataset. 
Table \ref{occlm_other} shows the comparison of our methods with PoseCNN \cite{YuXiang2017PoseCNNAC}, Pix2Pose \cite{KiruPark2019Pix2PosePC}, PVNet \cite{peng2019pvnet}, DPOD \cite{SergeyZakharov2019DPOD6P}, Hu et al. \cite{YinlinHu2020SingleStage6O}, HybridPose \cite{ChenSong2020HybridPose6O}, PVN3D \cite{he2020pvn3d} and FFB6D \cite{he2021ffb6d} in terms of the ADD(-S)-0.1d metric.
Our method achieves the state-of-the-art performance. Although the measurement results of eggbox and holepuncher are lower than PVN3D and FFB6D, but the results of other objects achieve the best performance among all methods. Our model exceeds FFB6D by 3.3\%. In particular, our method outperforms other methods by a margin of 11.4\% in the ADD(-S)-0.1d metric of the cat. The improved performance demonstrates that the proposed method is highly robust to occlusion. 
}


As for FFB6D \cite{he2021ffb6d}, we re-train its source code under the same settings suggested by the authors for fair comparisons. We can see that FFB6D performs very well, ranking the second place. 



\subsection{Visualization}
We demonstrate some visualization results from the YCB-Video dataset and the LineMOD dataset in Fig. \ref{vis} and Fig. \ref{lmvis}, respectively. 
We find in Fig. \ref{vis} that the ground truth of the object boxed in orange (see the right-bottom image) is not accurate, since the YCB-Video dataset is sampled from a fast-moving video, which produces some motion-blurred frames. 
The results show our method outperforms its competitor, even in the occlusion scenes, e.g., the object in the yellow box (see Figure \ref{vis} Column 2 and Column 4). 


\section{Conclusion}
In this paper, we propose a novel method, called SO(3)-Pose, for 6D instance-level object pose estimation. 
We verify that leveraging SO(3)-equivariance features to learn the keypoint offset and leveraging SO(3)-invariance features to learn the semantic label are beneficial to the pose estimation task. We propose a new loss function, namely SO3loss, to train the proposed network smoothly. Extensive experiments on two standard pose estimation datasets demonstrate the effectiveness of our proposed method, and show that it outperforms state-of-the-arts. 
However, our method is limited by the simple MLP structure.
In future, we will explore 
the powerful structures satisfying the SO(3)-equivariance property, such as transformer, to replace MLP. 


\bibliographystyle{eg-alpha-doi} 
\bibliography{egbibsample}       



\end{document}